\documentclass{article}

\usepackage[nonatbib,final]{bdl_2018}

\usepackage[utf8]{inputenc} 
\usepackage[T1]{fontenc}    
\usepackage{hyperref}       
\usepackage{url}            
\usepackage{booktabs}       
\usepackage{amsfonts}       
\usepackage{nicefrac}       
\usepackage{microtype}      
\usepackage{float}
\usepackage{graphicx}
\graphicspath{{../figs/}} 
\usepackage{amsmath}
\usepackage{amsthm}
\usepackage{isomath}
\usepackage{mdwlist}
\usepackage{wrapfig}
\usepackage{amsmath,amssymb,amsthm}
\usepackage{amsfonts}
\usepackage{amsopn}
\usepackage{epsfig}
\usepackage{ifthen}
\usepackage{algorithm}
\usepackage{algorithmic}
\usepackage{float}
\usepackage{array}
\usepackage{multirow}
\usepackage{makecell}
\usepackage{bm}
\usepackage{array,colortbl,multirow,multicol,booktabs}
\usepackage{caption}
\usepackage{subcaption}
\usepackage[square,sort,comma,numbers]{natbib}

\newcommand{\obs}{z}

\newcommand{\ft}{x}

\newcommand{\R}{{\rm I\!R}}

\title{Deep Factors with Gaussian Processes for Forecasting}

\author{
  Danielle C.~Maddix \\
  Amazon Web Services\\
  Palo Alto, CA 94303 \\
  \texttt{dmmaddix@amazon.com} \\
   \And
   Yuyang Wang \\
  Amazon Web Services\\
  Palo Alto, CA 94303 \\
     \texttt{yuyawang@amazon.com} \\
   \And
   Alex Smola \\
  Amazon Web Services\\
  Palo Alto, CA 94303 \\
   \texttt{smola@amazon.com} \\
}

\begin{document}

\maketitle

\begin{abstract}
  A large collection of time series poses significant challenges for classical and
  neural forecasting approaches.  
  Classical time series models fail
  to fit data well and to scale to large problems, but succeed at providing
  uncertainty estimates. 
  The converse is true for deep neural networks.  In this
  paper, we propose a hybrid model that incorporates the benefits of both
  approaches. Our new method is data-driven and scalable via a latent, global,
  deep component.  It also handles uncertainty through a local classical Gaussian Process model.  Our experiments demonstrate that our method obtains higher accuracy than state-of-the-art methods. 
\end{abstract}

\section{Introduction}


Some prevalent forecasting methods in statistics and econometrics have
been developed for forecasting individual or small groups of time
series.  These methods consist of complex models designed and tuned by
domain experts~\citep{harvey1990forecasting}. Recently, there has been
a paradigm shift from model-based to fully-automated data-driven
approaches.  This shift can be attributed to the availability of large
and diverse time series datasets in a wide variety of fields \citep{seeger2016bayesian}.  A substantial amount of data consisting of past behavior of related
time series can be leveraged for making a forecast for an individual
time series.  Use
of data from related time series allows for fitting of more complex
and potentially more accurate models without
overfitting. 



Classical time series methods, such as Autoregressive Integrated
Moving Average (ARIMA)~\citep{brockwell2013time}, exponential
smoothing~\citep{hyndman2008forecasting} and general Bayesian time
series~\citep{barber2011bayesian}, excel at modeling the complex
dynamics of individual time series of sufficiently long history.
These methods are computationally efficient, e.g.\ via a Kalman
filter, and provide uncertainty estimates.  Uncertainty estimates are
critical for optimal downstream decision making.  These methods are
local, that is, they learn one model per time series.  As a
consequence, they cannot effectively extract information across
multiple time series. These classical methods also have challenges
with cold-start problems, where more time series are added or
removed over time. 

Deep neural networks (DNNs), in particular, recurrent neural networks (RNNs),
such as LSTMs \citep{hochreiter1997long} have been successful
in time series forecasting~\citep{flunkert2017deepar,wen2017multi}.
DNNs are generally effective at extracting patterns across multiple
time series.  Without a combination with probabilistic methods, such as variational dropout \cite{gal2016} and deep Kalman filters \cite{krishnan2015deep}, DNNs can be prone to overfitting and have challenges in modeling uncertainty \citep{garnelo2018conditional}.  

The combination of probabilistic graphical models with deep neural networks has been an active research area recently~\cite{krishnan2015deep,krishnan2017structured,fraccaro2016sequential,fraccaro2017disentangled}. In the time series forecasting domain, a recent example is~\cite{rangapuram2018}, where the authors combine RNNs and State-Space Models (SSM) for scalable time series forecasting. Our work in this paper follows a similar theme: we propose a novel and scalable global-local method, Deep Factors with Gaussian Processes.  
It is based on a global DNN backbone and
local Gaussian Process (GP)  model for computational efficiency.
The global-local structure extracts complex non-linear
patterns globally while capturing individual random effects for each
time series locally.  The main idea of our approach is to represent each time series as a combination of a global
time series and a corresponding local model. The global part is given
by a linear combination of a set of deep dynamic factors, where the
loading is temporally determined by attentions. The local model is
a stochastic Gaussian Process (GP), which allows for the uncertainty to propagate forward in
time.  

\section{Deep Factor Model with Gaussian Processes}
\label{sec:model}

We first define the forecasting problem that we are aiming to solve.  Let $\mathcal{X}\subset \R^d$ denote the input features space and $\mathcal{Z}\subset \R^k$ the space of the observations. We are given a set of $N$ time series with the $i^{\text{th}}$ time series consisting of $(\ft_{i,t}, \obs_{i,t})\in \mathcal{X} \times \mathcal{Z}, t=1, \cdots, T,$ where $\ft_{i,t}$ are the input co-variates, and $\obs_{i,t}$ is the corresponding observation at time $t$. Given a forecast horizon $\tau \in \mathbb{N}^+$, our goal is to calculate the joint predictive distribution of future observations, 
 \[
 p(\{\obs_{i, T+1:T+\tau}\}_{i=1}^N | \{\ft_{i, T+1:T+\tau}\}_{i=1}^N, \{\mathcal{D}_i\}_{i=1}^N), 
 \]
 where $\mathcal{D}_i = \{(\ft_{i}, \obs_{i})\}$ denotes the $i^{\text{th}}$ time series with corresponding features. For concreteness, we restrict ourselves to univariate time series ($k=1$). 

\subsection{Generative Model}
\label{sect:genmodel}

We assume that each time series $\obs_i$ is governed by the following two components: fixed and random.

\begin{wrapfigure}{L}{4cm}
\centering
\includegraphics[width=.25\textwidth]{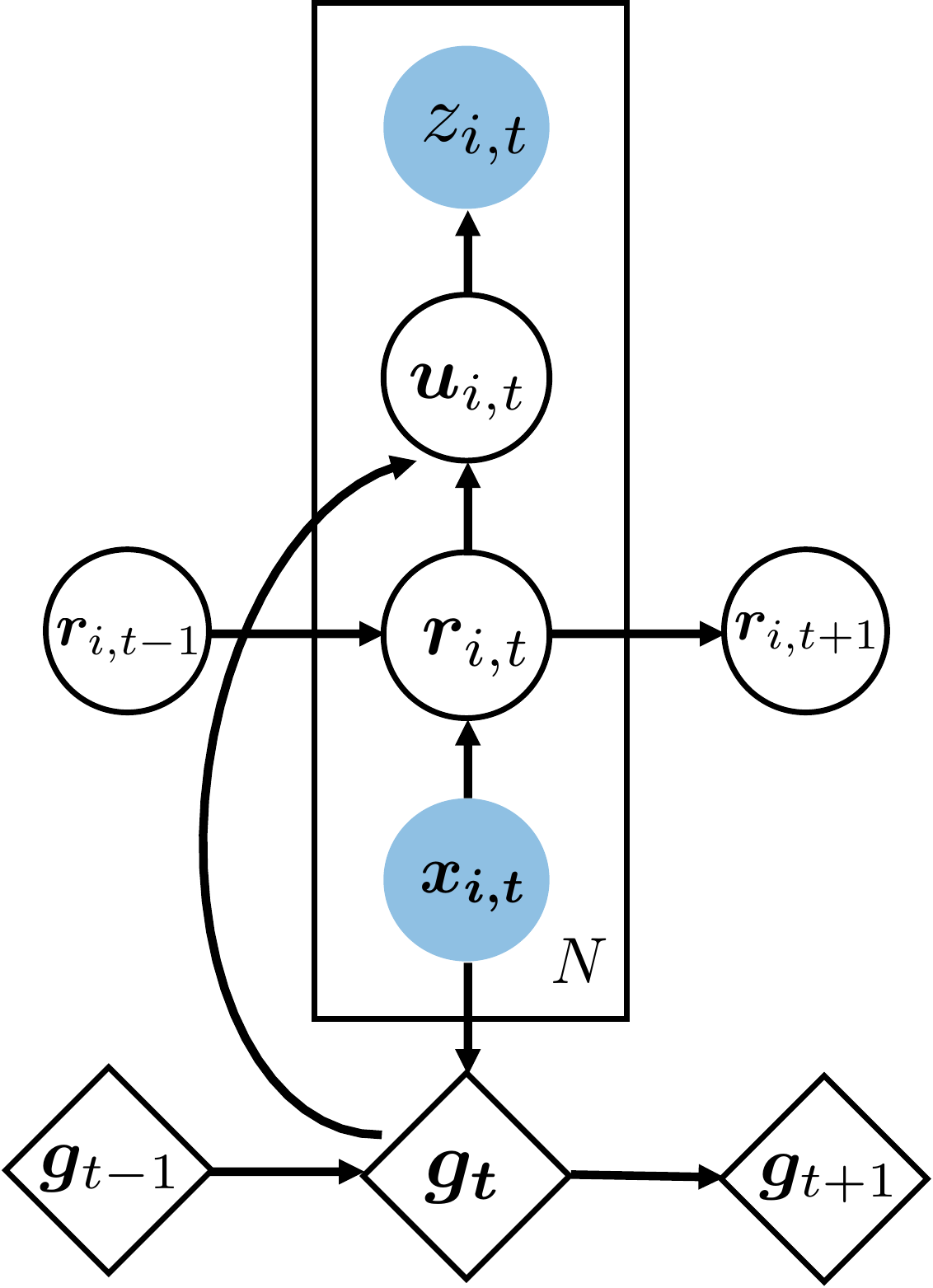}
\caption{Plate graph of the proposed Deep Factors with Gaussian Processes model. The diamond nodes represent deterministic states.}
  \label{fig:platemodel}
  \vspace{-1cm}
\end{wrapfigure}
Fixed effects are common patterns that are given by linear combinations of $K$ latent 
  global deep factors, $g_{k,t}$.  These deep factors can be thought of as dynamic principal components
  or eigen time series that drive the underlying dynamics of all the time series.  

Random effects, $r_{i}$, are the local fluctuations that are chosen to be the Gaussian Process~\citep{rasmussen2006gaussian}, i.e., $r_i \sim \textbf{GP}(0, \mathcal{K}_i(\cdot, \cdot))$, where the covariance $\mathcal{K}_i$ is a kernel matrix and $r_{i,t} = r_i(x_{i,t})$.  

The observed value $\obs_{i,t}$ at time $t$, or more generally, its latent function $u_{i,t}$ such 
that $\obs_{i,t} \sim p(\cdot|u_{i,t}(x_{i,t}))$, can be expressed as a sum of
the weighted average of the global patterns and its local fluctuations. The summary of this generative model is given in Eqn. \eqref{eqn:df_gp}, and is illustrated in Figure \ref{fig:platemodel}.  For simplicity, we consider $w_i(x_{i,t}) := w_i$ to be the embedding of time series $i.$
\begin{equation}
\begin{split}
\text{random effect}:\quad &r_{i} \sim \textbf{GP}(0, \mathcal{K}_i(\cdot, \cdot)), \\
\text{fixed effect}:\quad  &f_{i,t} = w_i^\top g_t(x_{i,t}),\\
\text{emission}: \quad & z_{i,t} \sim p(\cdot|u_{i,t}),\ u_{i,t} = f_{i,t} + r_{i,t}.
\label{eqn:df_gp}
\end{split}
\end{equation}  

We use a global dynamics factors RNN or a set of $K$ univariate-valued RNNs to generate $g_t \in \R^K$. The RNNs are learned globally to capture the common patterns from all time series.  For each time series at time $t$, we use attention networks to assign stationary attentions $w_{i} \in\R^K$ to the dynamic factors $g_t$.  This determines the group of the global factors to focus on and the  relevant segment of histories. At a high level, the weighting gives temporal attention to different global factors.

\subsection{Inference and Learning}
Given a set of $N$ time series generated by Eqn. \eqref{eqn:df_gp}, our goal is to estimate $\mathbf{\Theta}$, the parameters in the global RNNs, attention network and the hyperparameters in the kernel function. To do so, we use maximum likelihood estimation, where
$
\mathbf{\Theta} = \text{argmax} \sum_i\log p(z_i),
$
Computing the marginal likelihood may require doing inference over the latent variables.  In our case, $p(\cdot|u_{i,t})$ is Gaussian, and the marginal likelihood can be computed easily as,
\[
p(z_i) = \mathcal{N}(f_i, \mathcal{K}_i + \sigma_i^2\mathbb{I}).
\]
For non-Gaussian likelihoods, classical techniques, such as Box-Cox transform \cite{box2015time} or variational inference in the framework of Variational Auto Encoder (VAE)~\cite{kingma2013auto,rezende2014stochastic}, can be used. This is a direction of future work.

\section{Experiments}
\label{sec:exp}
The model is implemented in MXNet Gluon~\citep{chen2015mxnet} with a RBF kernel \cite{gardner2018} using the \texttt{mxnet.linalg} library \cite{seegar2017,dai2018}. We use a p3.4xlarge SageMaker instance in all our experiments.  The global factor network is chosen to be LSTM with 1 hidden layer and 50 hidden units. We fix the number of factors to be 10. 

\begin{figure}[H]
\begin{subfigure}{0.5\textwidth}
  \includegraphics[width=\linewidth]{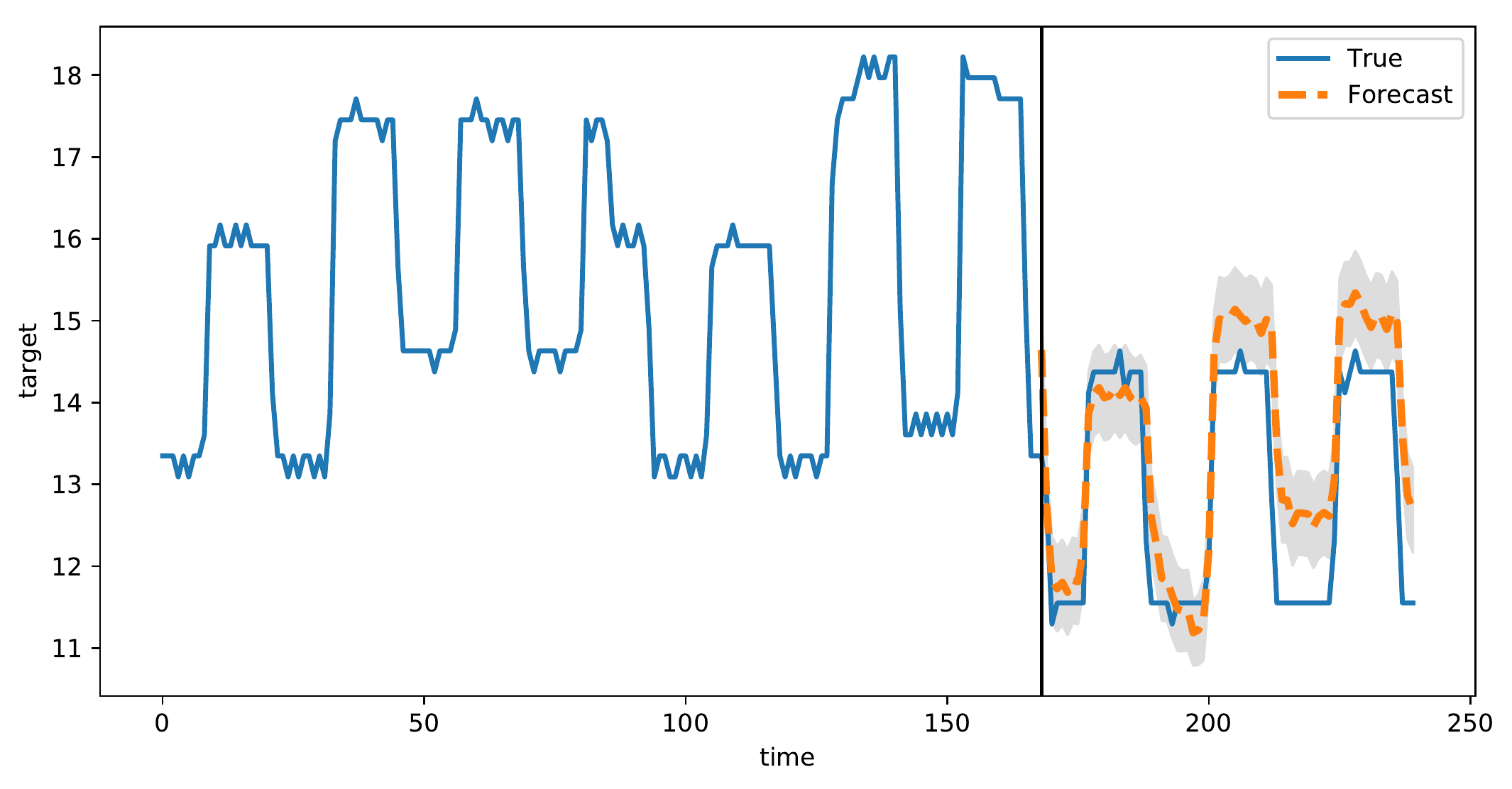}
  \caption{ \texttt{electricity}}
  \label{elect}
\end{subfigure}%
\begin{subfigure}{0.5\textwidth}
  \includegraphics[width=\linewidth]{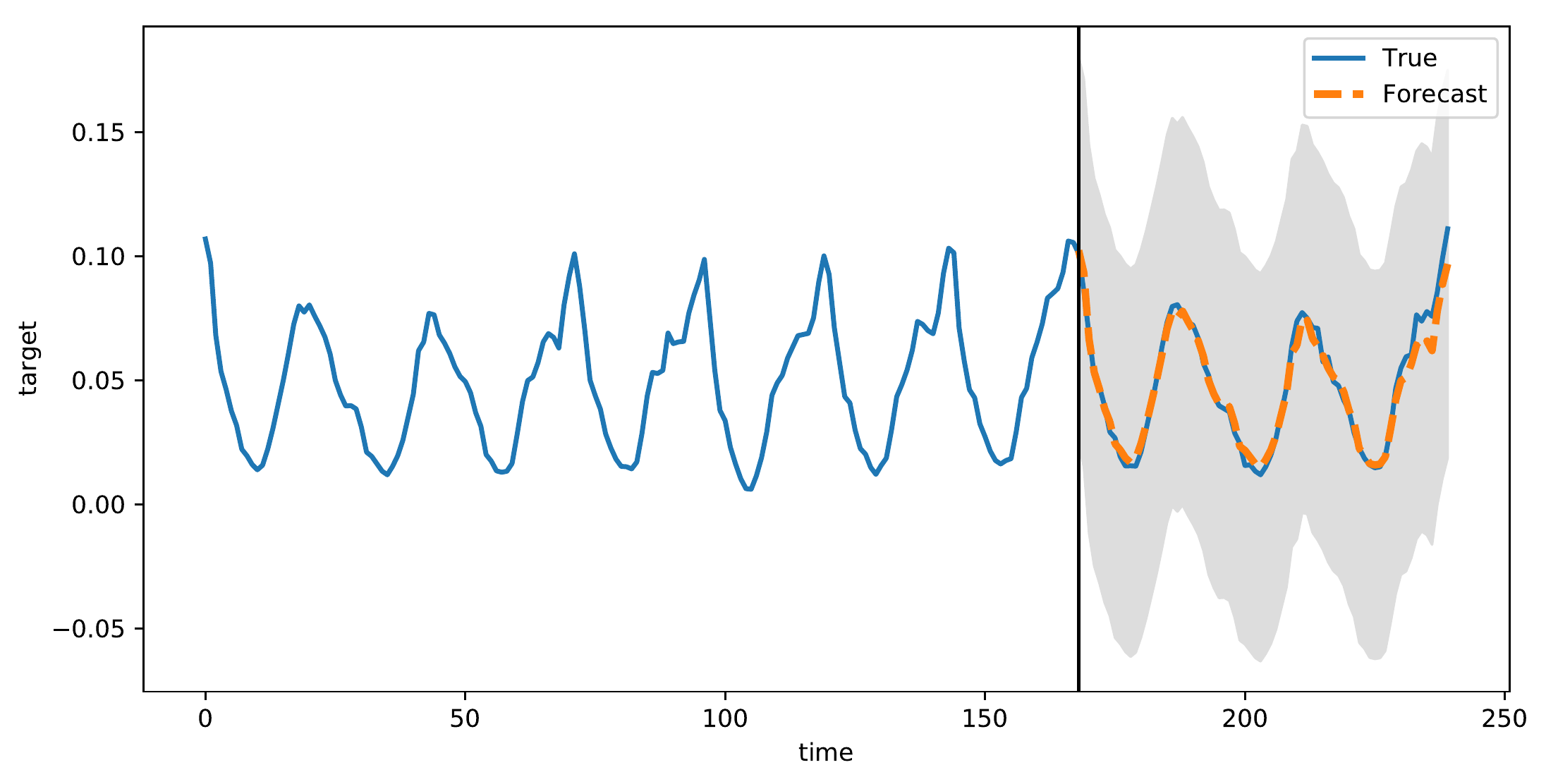}
  \caption{\texttt{traffic}}
  \label{traffic}
\end{subfigure}
 \caption{The dashed orange curve shows the forecast of the proposed global LSTM with GP local model.  The black vertical line marks the division between the training and prediction regions.}
\label{fig:GP_res}
\end{figure}

To assess the quality of the proposed model, we limit the training, sometimes artificially by pruning the data, to only one week of time series.  This results in 168 observations per time series. Figures \ref{elect}-\ref{traffic} show that the forecasts qualitatively on the publicly available datasets \texttt{electricity} and \texttt{traffic} from the UCI data set~\citep{Dua:2017,yu2016temporal}.  

We use the quantile loss to evaluate the 
probabilistic forecast. For a given quantile $\rho\in(0,1)$, a target value $\obs_t$ and $\rho$-quantile prediction $\widehat{\obs}_t(\rho)$, the $\rho$-quantile 
loss is defined as
\begin{align*}
\text{QL}_\rho[\obs_t, \widehat{\obs}_t(\rho)] &= 2\big[\rho(\obs_t - \widehat{\obs}_t(\rho))\mathbb{I}_{\obs_t - \widehat{\obs}_t(\rho) > 0} \nonumber + (1-\rho)(\widehat{\obs}_t(\rho) - \obs_t)\mathbb{I}_{\obs_t - \widehat{\obs}_t(\rho) \leqslant 0}\big].
\end{align*}
We use a normalized sum of quantile
losses, 
$\sum_{i,t} \text{QL}_\rho[\obs_{i,t}, \widehat{\obs}_{i,t}(\rho)] / \sum_{i,t} |\obs_{i,t}|,$
to compute the quantile losses for a given span across all time series.
We include results for $\rho=0.5, 0.9,$ which we abbreviate as the P50QL (mean absolute percentage error (MAPE)) and P90QL, respectively. 
We also report the root mean square error (RMSE), which is the square 
root of the aggregated squared error normalized by the product of number of 
time series and the length of the time series in the evaluation segment. 

Table \ref{tab:results-mu} compares with DeepAR (DA), a state-of-art 
RNN-based forecasting algorithm on the publicly available AWS SageMaker ~\citep{flunkert2017deepar, janu2018} and Prophet (P), a Bayesian structural time series model~\citep{taylor2017forecasting}. To ensure a fair comparison, we set DeepAR to have the same 1-layer 50 hidden units network configuration, with the number of epochs set to be 2000. The results show that our model outperforms the others, in particular with respect to the P90 quantile loss.  This shows that we are better at capturing uncertainty.  

\begin{table}[H]
\centering
\footnotesize{
\begin{tabular}{l|c|ccc|ccc|rrr}
\toprule
\multirow{2}{*}{\textsc{ds}} & \multirow{2}{*}{\textsc{hrzn}} & \multicolumn{3}{c}{\textsc{p50ql}} & \multicolumn{3}{c}{\textsc{p90ql}} & \multicolumn{3}{c}{\textsc{RMSE}}\\
\cline{3-11}
{} & & DA & P & DFGP  & DA & P & DFGP & DA & P & DFGP\\
\midrule 
\multirow{2}{*}{\texttt{elec}} & 3d & 0.216 & 0.149 &\textbf{0.109}  &  0.182 & 0.103 & \textbf{0.061} & 1194.421 & 902.724 & \textbf{745.175}  \\
 & 24hr &  0.132 & 0.124 &\textbf{0.103}  &0.100   & 0.091 &\textbf{0.074}  & 2100.927 & 783.598 &\textbf{454.307} \\
\midrule
\multirow{2}{*}{\texttt{traf}} & 3d & 0.348 & 0.457 &\textbf{0.137} &  0.162 & 0.207 & \textbf{0.093} &0.028 & 0.032 & \textbf{0.021}\\
& 24hr & 0.268 & 0.380 &\textbf{0.131} & 0.149  & 0.191 & \textbf{0.090} & 0.024 & 0.028 & \textbf{0.019}\\
\bottomrule
\end{tabular}}
\vspace{.1cm}
\caption{Results for short-term (3-day forecast) and near-term (24-hour forecast) scenario with one week of training data on \texttt{electricity, traffic}.}
\label{tab:results-mu}
\end{table}

\section{Conclusion}
We propose a novel global-local model, Deep Factors with Gaussian Processes, for forecasting a collection of related time series. Our method differs from other 
global-local models by combining classical Bayesian probabilistic models with deep learning techniques that scale.  
We show promising experiments that demonstrate the effectiveness and potential of our method in learning across multi-time series and propagating uncertainty.  

\bibliographystyle{unsrt}
\small
\bibliography{../dssm,../ts}

\onecolumn
\appendix

\end{document}